\documentclass{article}
\PassOptionsToPackage{numbers, compress}{natbib}

\usepackage[preprint]{nips_2018}

\usepackage[utf8]{inputenc} 
\usepackage[T1]{fontenc}    
\usepackage{hyperref}       
\usepackage{url}            
\usepackage{booktabs}       
\usepackage{amsfonts}       
\usepackage{nicefrac}       
\usepackage{microtype}      

\title{Transferring Deep Reinforcement Learning with Adversarial Objective and Augmentation}

\author{
    Shu-Hsuan~Hsu, I-Chao Shen, Bing-Yu Chen
    \\
    Department of Computer Science and Information Engineering\\
    National Taiwan University\\
    \texttt{{ssarcandy,jdily}@cmlab.csie.ntu.edu.tw; robin@ntu.edu.tw} \\
}
\newboolean{revising} 

\usepackage{amsmath}
\usepackage{amssymb} 
\usepackage{amsfonts}
\usepackage{multirow}
\usepackage{subcaption}
\usepackage{color}
\usepackage{ifthen}
\usepackage{xcolor}
\usepackage[normalem]{ulem}
\usepackage{enumitem}
\usepackage[encapsulated]{CJK} 
\usepackage{float}
\usepackage{graphicx}

\setboolean{revising}{false}

\newcommand{\ie}{i.e.}

\newcommand{\figname}{Figure}

\newcommand{\secname}{Section}

\newcommand{\etal}{{\textit{et~al.}}}

\ifthenelse{\boolean{revising}}
{
\newcommand{\ichao}[1]{\textcolor{blue}{#1}}
\newcommand{\iccmt}[1]{\textcolor{purple}{\textbf{ichao:} #1}}
\newcommand{\ssarcandy}[1]{\textcolor{orange}{#1}}
\newcommand{\robin}[1]{\textcolor{magenta}{#1}}
\newcommand{\ignore}[1]{}
}{
\newcommand{\ichao}[1]{#1}
\newcommand{\iccmt}[1]{}
\newcommand{\ssarcandy}[1]{#1}
\newcommand{\robin}[1]{#1}
\newcommand{\ignore}[1]{}
}

\begin{document}

\maketitle


\begin{abstract}
In the past few years, deep reinforcement learning has been proven 
\robin{to} solve \ssarcandy{problems \robin{which} have complex states like video games or board games}. The next step of intelligent agents would be able to generalize between tasks, \robin{and} using prior experience to pick up new skills more quickly. However, most reinforcement learning algorithms for now are often suffering from catastrophic forgetting even when facing a very similar target task.
Our approach enables \robin{the} agent\robin{s to} generalize knowledge from a single source task, and boost the learning progress with a semi-supervised learning method when facing \robin{a} new task.
We evaluate this approach on Atari games, \robin{which is} a popular reinforcement learning benchmark, and show that it outperforms common baselines based on pre\robin{-}training and fine\robin{-}tuning.
\end{abstract}

\section{Introduction}
\label{sec:introduction}

Deep Reinforcement Learning (DRL), the combination of reinforcement learning methods and deep neural network function approximators, has recently shown considerable success in \ssarcandy{challenging tasks that have very complex states and many available actions, such as arcade video games~\cite{mnih2015human}, robotic manipulation~\cite{levine2016end}, and even the challenging classic games \robin{-} Go~\cite{silver2016mastering}}. 
These methods can learn features that \robin{are} often better than hand-craft \robin{one}s\robin{, which} require more domain knowledge. For example, 
Deep Q-Network (DQN) \cite{mnih2015human} \robin{is} one of the most famous DRL method\robin{s}, \robin{and} has achieved super human performance on the Arcade Learning Environment (ALE) \cite{bellemare13arcade}, \robin{which is} a benchmark of Atari 2600 arcade games. 

Although the DRL algorithms can usually learn how to take the best action based on the state of the environment, but it can only learn a single environment at a time, despite the existence of similarities between those environments. 
For example, the tennis-like game of pong and the squash-like game of breakout are both similar in that each game consists of trying to hit a moving ball with a rectangular paddle, but an agent that is good at pong cannot handle breakout well, \ichao{and vice versa.} 
Another issue of DRL is that training DRL agents can be very time-consuming, \robin{so} many researcher\robin{s are} stud\robin{ying} on the methods that can speed up \robin{the} training time \cite{mnih2016asynchronous,van2016deep}.

Some \robin{methods} speed up \robin{the} learning on new tasks by perform\robin{ing} cross environment transfer \cite{rusu2016progressive,parisotto2015actor}, but they all need to pre-train an agent on multiple source environments to generalize the knowledge, which is very time\robin{-}consuming.
In this work, we \robin{are} trying to leverage the prior knowledge learn\robin{ed} by a\robin{n agent in a} single source environment to speed up \robin{the} agent to handle \robin{another} new environment\robin{. U}sing \robin{the prior knowledge in} only one source environment can minimize the \robin{training} time \robin{in another} new environment, and can also solve some issues of reinforcement learning, including unable to handle similar tasks and long training time problems.

\robin{In this paper, w}e propose a semi-supervised \ssarcandy{transfer learning} method that uses the concept of the \ssarcandy{adversarial objective}. More specifically, 
a mapping \robin{function is learned} from \robin{the} target observations to \robin{the} source feature space by fooling a domain discriminator that tries to distinguish the encoded target observation from \robin{the} source examples. 
We also found it is helpful for transfer\robin{ring} knowledge by \ssarcandy{altering the visual content with proper setting when training on \robin{the} source task, that is, adding proper augmentation when training will be helpful on transferring knowledge to another task.} While our approach can integrate into any DRL algorithm, we show \robin{our} results 
by combining it with DQN \cite{mnih2015human} algorithm and \robin{performing on} 
Atari 2600 
\robin{benchmark}.

    
Our contributions are two-fold\robin{s}:
First, we propose a method that can leverage \robin{the} knowledge \robin{learned} from a single source task 
to 
speed up \robin{the} training on \robin{another} new target environment. Second, we found that perform\robin{ing} proper augmentation on the environment to train a source agent and use it as \robin{a} target task initialization often can help. 
With these proposed methods, the overall learning on \robin{the} target task \robin{can be} accelerated compar\robin{ing} with \robin{the} baselines that we have considered.

The rest of this paper \robin{is} organized as follows. In \secname~\ref{sec:related_work}, we surveyed several previous methods for reinforcement learning, domain adaption, and multi-task agent. \robin{In} \secname~\ref{sec:background}\robin{,} we detailed the knowledge of traditional Q-learning and deep Q network, as it is the most popular reinforcement learning algorithm and is used in our method. \robin{In} \secname~\ref{sec:approach}\robin{,} 
the two main methods used in our approach \robin{are described}, including \textit{adversarial objective} and \textit{augmentation}. \robin{In} \secname~\ref{sec:experiments}\robin{,} 
a series of experiments \robin{is performed} to evaluate our method 
\robin{with} a discussion in detail\robin{s}. Finally, \secname~\ref{sec:conclusion} concludes the paper.

\section{Related Work}
\label{sec:related_work}
\subsection{Reinforcement Learning}
Reinforcement learning is a method that can learn how to map \ichao{observations}\ignore{situations} to actions \ichao{by}\ignore{try to} maximiz\ichao{ing} the reward~\cite{sutton1998reinforcement}. 
Unlike \ichao{supervised learning paradigm}\ignore{other machine learning}\ignore{will} usually \ichao{requires}\ignore{provide} correct labels \ichao{in order to learn a new task}\ignore{ used for direct feedback}, the reinforcement learner is not told \ichao{the best }\ignore{which }action\ignore{ is best}, but instead \ichao{acting according to the best reward}.\ignore{ need to discover which actions yield the most reward by trying them. }
The chosen action \ichao{affects}\ignore{taken by the agent may affect not only} the immediate reward,\ignore{but also} the next \ichao{observation}\ignore{situation}, and\ignore{ through that,} all subsequent rewards. 

In recent years, \ichao{with}\ignore{thanks to} the significant \ichao{advances of}\ignore{progress in} deep learning, \ichao{especially with convolutional neural network}, numerous researchers have attempted to use deep learning to solve the reinforcement learning tasks \cite{mnih2015human,mnih2016asynchronous,van2016deep,mnih2013playing}. 
Mnih~\etal~\cite{mnih2015human} presented Deep Q-Network (DQN) that uses a deep network to approximate the state-action value function \ichao{used in traditional Q-learning~\cite{watkins1992q}}.
\ichao{It} solves the storing space issue in traditional state-action tabular representation (Q-table) \cite{watkins1992q}. 
\iccmt{I don't understand..}
Van Hasselt \etal~\cite{van2016deep} further extend\ichao{ed} the DQN by solved an overestimate action values issue suffered for the DQN algorithm and leads to much better performance. 
Mnih~\etal~\cite{mnih2016asynchronous} introduced a deep reinforcement learning framework that uses asynchronous gradient descent for \ichao{optimizing}\ignore{optimization of} deep neural network controllers, that achieve state-of-the-art performance\ignore{ in training time}. 
Fortunato~\etal~introduced a method called \ichao{NoisyNet}\ignore{NosiyNet}~\cite{DBLP:journals/corr/FortunatoAPMOGM17}, \ssarcandy{by added parametric noise to its weight to induce stochasticity of the agent's policy, that can be used to aid efficient exploration.
Replacing the conventional exploration heuristics for A3C, DQN with NoisyNet can yield higher scores for a wide range of Atari games.}

\subsection{Domain Adaptation}
Deep neural networks are able to learn powerful representations from large quantities of labeled input data, however, they usually fail to \ichao{generalize to}\ignore{handle when the input} \ichao{new datasets}\ignore{distributions} \ichao{with even minor}\ignore{have some} changes.
\ichao{In recent years,} numerous\ignore{ studies on} domain adaption methods have been proposed~\cite{tzeng2017adversarial,sun2016deep,ganin2015unsupervised,sun2016return,xiao2016learning}. 
Instead of collecting labeled data and training a new classifier for every possible scenario, unsupervised domain adaptation methods are trying to compensate for the degradation in performance by transferring knowledge from labeled source domains to unlabelled target domains. 
Sun~\etal~\cite{sun2016return} proposed \textit{CORAL}, that can align the second-order statistics of \ssarcandy{the source and target distributions of features with a linear transformation. }
Sun and Saenko~\cite{sun2016deep} \ichao{proposed a differentiable \textit{CORAL} loss} that \ichao{can be minimized end-to-end and be integrated with general deep neural networks.}
Ganin and Lempitsky~\cite{ganin2015unsupervised} use\ichao{d} a domain classifier with a gradient reversal layer that multiplies the gradient by a negative constant, which can ensure that the feature distributions over the two domains are made similar, thus resulting in the domain-invariant features.
Tzeng~\etal~\cite{tzeng2017adversarial} present a method that uses the concept of \ignore{Generative }\ichao{adversarial learning}\ignore{Network} \cite{goodfellow2014generative}; 
\ichao{they} pre-train a source encoder CNN with labeled data, then perform adversarial adaptation by learning a target encoder CNN such \ichao{in order to fool the }\ignore{that a} discriminator.  
\ichao{This} result\ichao{s}\ignore{ing} in a target encoder CNN that can produce features that similar to source features.

\subsection{Multi-task Agent}
Although \ichao{an agent trained by} DRL can \ichao{outperform} human-expert level across many Atari games, but \ichao{fail to generalize across multiple games}\ichao{each agent can only play a single game. }
\iccmt{What is ``model compression'' technique here..?}
To \ichao{address}\ignore{solve} this issue, some researchers attempt to train a single agent to handle multiple tasks by integrate model compression technique to deep reinforcement learning.
Parisotto~\etal~\cite{parisotto2015actor} first present\ichao{ed} ``Actor-Mimic'', exploit\ichao{ed} the use of \ichao{DRL}\ignore{deep reinforcement learning} and model compression techniques to train a single policy network that learns how to act in a set of distinct tasks by using the guidance of several expert teachers. 
Yin and Pan~\cite{yin2017knowledge} propose\ichao{d} a new multi-task policy distillation architecture \ichao{by} concatenat\ichao{ing} a set of task-specific convolutional layers and shared multi-task fully connected layers.
The shared multi-task fully connected layers enable the agent to learn a generalized \ichao{policy}\ignore{reasoning about when to issue what action under different circumstances}.
Sharma and Ravindran~\cite{sharma2017online} propose\ichao{d} a\ignore{simple yet efficient} multi-task learning framework which solves multiple\ignore{goal-directed} tasks in an online  learning setup without the need \ichao{of training multiple task-specific teacher (expert) networks.}

\section{Background}
\label{sec:background}
Our method is aiming for leveraging knowledge \ichao{learned} from a source task to help speed up \ichao{learning}\ignore{training} on \ichao{a} new target task.
In this work, we choose to integrate the proposed method into deep Q networks (DQN)~\cite{mnih2015human}, since DQN is one of the most popular deep reinforcement architecture. 
In this section, we detailed the knowledge of Q-learning in \secname~\ref{subsec:q-learning} and further describe DQN in \secname~\ref{subsec:dqn}.

\subsection{Q-Learning}
\label{subsec:q-learning}
Reinforcement learning involves an agent, a set of states $S$, and a set of actions $A$. By performing an action $a \in A$, the agent transitions from state $(s)$ to another state $({s}')$.
After performing an action, \ichao{the agent} will get a reward. 
The goal of reinforcement learning agent is to maximize its total reward.

Q-learning was first introduced by Watkin \etal~\cite{watkins1992q}, a reinforcement learning method that can estimate the optimal action value by interacting with the environment. 
The Q \ichao{stands for}\ignore{means} the quality of a ``state-action pair'', by interacting with environment and record all the state-action transitions, it can gradually update the Q-value of the state-action pair. 
The quality of a state-action pair is defined as:
\begin{equation}
Q(s, a) \longrightarrow  \mathbb{R}    
\end{equation}

For each step $t$, the agent select\ichao{s} action $a_{t}$ and transits from state $s_{t}$ to state $s_{t+1}$. 
The new Q-value is update\ichao{d} as the last step Q-value $Q(s_{t-1}, a_{t-1})$ plus the current reward $r_{t}$ and estimate future optimal value multiply by discount factor $r_{t}+\gamma \max_{a}Q(s_{t+1}, a)$, and the update equation then can describe as follow:

\begin{equation}
Q(s_{t}, a_{t}) = (1-\alpha)Q(s_{t-1}, a_{t-1}) + \alpha(r_{t}+\gamma \max_{a}Q(s_{t+1}, a))
\end{equation}
where $\alpha$ is learning rate and $\gamma \in [0, 1]$ is a discount factor that trades off the importance of immediate and \ichao{future}\ignore{later} rewards.

\iccmt{
The writing is a bit messy...
Will come back later.
}

\subsection{Deep Q Network}
\label{subsec:dqn}
Although Q-learning can be used to estimate optimal \ichao{policy}\ignore{action}, instead of learning all action values in all states, which may be too large to learn in complex environments, we can learn a parameterized value function $Q(s, a; \theta)$, where $\theta$ are the parameters of the network. 
For updating the network parameters after taking action $a_{t}$ in state $s_{t}$ and observing the reward $r_{t+1}$ and next state $r_{t+1}$ is:

\begin{equation}
\theta_{t+1} = \theta_{t} + \alpha(Y_{t} - Q(s_{t}, a_{t}; \theta_{t}))\nabla_{\theta_{t}}Q(s_{t}, a_{t}; \theta_{t}) 
\end{equation}

and $Y_{t}$ is defined as:

\begin{equation}
Y_{t} = r_{t+1} + \gamma \max_{a}Q(s_{t+1}, a; \theta_{t})
\end{equation}

where $\alpha$ is learning rate and $\gamma \in [0, 1]$ is a discount factor that trades off the importance of immediate and later rewards.

A most famous deep reinforcement learning algorithm is deep Q network (DQN) that introduced by Mnih et al. \cite{mnih2015human}. DQN is a multi-layered neural network that for a given state outputs a vector of action values. For an $n$-dimensional state space and an action space containing $m$ actions, the neural network is a function from $\mathbb{R}^{n}$ to $\mathbb{R}^{m}$. 

There have two important components that used in DQN, a target network and replaying memory. The target network is same as the online network, and the parameters of target network $\theta^{'}$ will copy from the online network every $t$ steps, and kept fixed on all other steps. The target network is used to predict the true Q value when learning ($Y_{t}$). For the replaying memory, observed transitions are stored for some time and sampled uniformly from this memory buffer to update the network. Both the target network and the replaying memory improve the training stability and overall performance of the algorithm \cite{mnih2015human}. And to explore environment more stable and faster, DQN also using $\epsilon$-greedy\cite{sutton1998reinforcement} agent's policy, that agent will select a random action with certain possibilities.


In this work, we split the agent into two networks as shown in \figname~\ref{fig:3_training} (a),  a feature encoder $F$ and a Q-value predictor $V$. Feature encoder extract features based on the input observations, and the Q value predictor uses these encoded features to predict the corresponding action Q values.

\section{Approach}
\label{sec:approach}


In this section, we present a transfer learning \ichao{method}\ignore{architecture} that can speed up the learning progress when facing new target \ichao{task}\ignore{environment}.
Furthermore, we found a novel data augmentation method that can help single task agent to avoid over-fitting and thus learn more \ichao{generalized}\ignore{core} \ichao{policy}\ignore{features}.

\subsection{Transfer with Adversarial Objective (AdvTransfer)}
\label{subsec:transfer_learning}

We present a framework for unsupervised domain adaption between deep reinforcement learning tasks. 
Assume a source environment $Env_{s}$ and a target environment $Env_{t}$, there \ichao{are}\ignore{have some} domain shift\ichao{s} between $Env_{s}$ and $Env_{t}$. 
We have access to the reward and next state after performed an action on $Env_{s}$.

\begin{figure*}[t!]
\centering
\includegraphics[width=\linewidth]{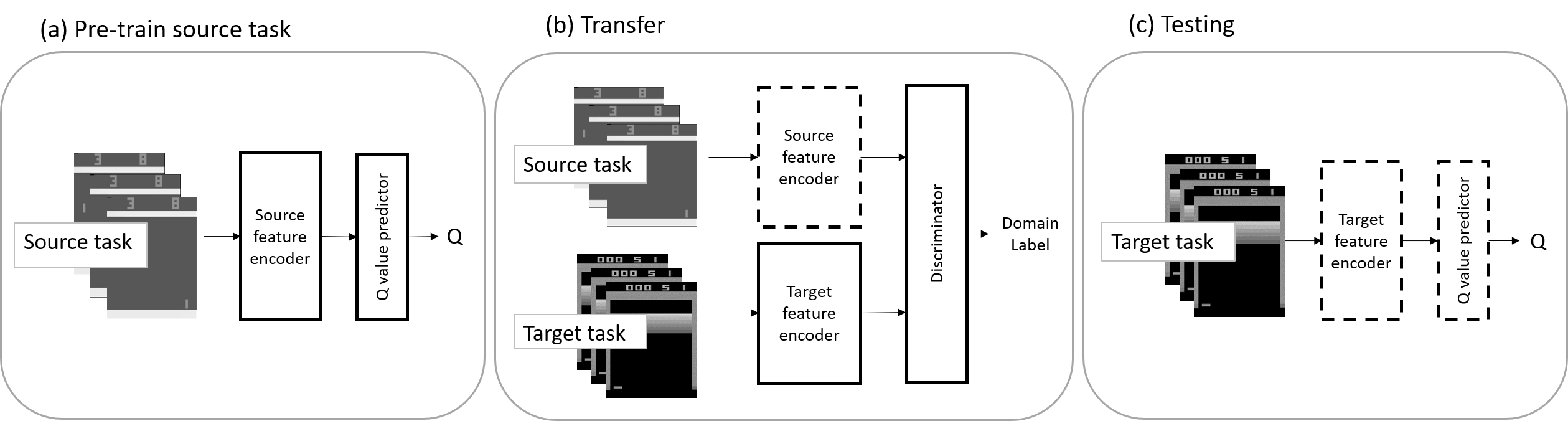} 
\caption{An overview of our transfer process. (a) We first well-trained an agent on source task environments, then (b) we train a domain classifier and target feature encoder that have an adversarial objective to learn a map that maps target feature encoder to source feature encoder. (c) At the testing time, we can use source task Q value predictor directly because the target feature encoder's output features are similar to source feature encoder. The dashed line means fixed parameters.}
\label{fig:3_training}
\end{figure*}

The overview is shown in \figname~\ref{fig:3_training}, we first pre-trained the source task agent on $Env_{S}$, training a feature encoder $F_{S}$ and Q-value predictor $V_{S}$ that can return a vector of Q values for all possible actions by given features of observation, thus can select the best action $a$.
And our goal is to learn a target feature encoder $F_{T}$ and Q-value predictor $V_{T}$ that can handle $Env_{T}$ as fast as possible.

We integrate generative adversarial network~\cite{goodfellow2014generative} concept into transfer progress, as shown in \figname~\ref{fig:3_training} (b). The target feature encoder $F_{T}$ plays the role of generator, and a domain classifier $D$ that can predict the domain label (source or target domain) by seeing the encoded features output by $F_{S}$ and $F_{T}$. 
We then perform adversarial adaption by train\ichao{ing} the $F_{T}$ \ichao{in order to} to \ichao{prevent}\ignore{fool} the domain classifier \ichao{from predicting the correct domain label from the encoder features.}
The target feature encoder (Generator) and domain classifier (Discriminator) are playing counterparts, and at the end of training, the target feature encoder $F_{T}$ will learn the feature representations that can be generalized across both source and target domain.


In the adversarial objective approach, the main goal is to regularize the learning of the source and target mappings, we can minimize the distance between $F_{S}(S_{s})$ and $F_{T}(S_{t})$ distributions, where $S_{s}$ and $S_{t}$ is the states of $Env_{S}$ and $Env_{T}$. 
If the distribution between $F_{S}(S_{s})$ and $F_{T}(S_{t})$ are similar, then we can directly apply source task Q-value predictor $V_{s}$ to the $F_{t}$, skipping the need to learn a $V_{t}$.

We first describe the domain classifier, $D$, which classifies
whether encoded features are drawn from the source or the target domain. Thus $D$ is optimized according to a standard cross entropy loss, $L_{D}(S_{s}, S_{t}, F_{s}, F_{t})$ where the labels indicate the origin domain, defined below:

\begin{equation}
    L_{D}(S_{s}, S_{t}, F_{s}, F_{t}) = -\log(D(F_{s}(S_{s}))) - \log(1 - D(F_{t}(S_{t})))
\end{equation}

And for the generator, we train with the standard loss function with inverted labels \cite{goodfellow2014generative}, then the loss can be described as:

\begin{equation}
    L_{G}(S_{t}, D) = -\log(D(F_{t}(S_{t})))
\end{equation}

Then, the source and target mappings are optimized according to an adversarial objective, they are optimized to confuse $D$ to unable to predict reliable domain label.

\subsection{Augmentation}
\label{subsec:augmentation}

\begin{figure}[ht]
\centering
\includegraphics[width=\linewidth]{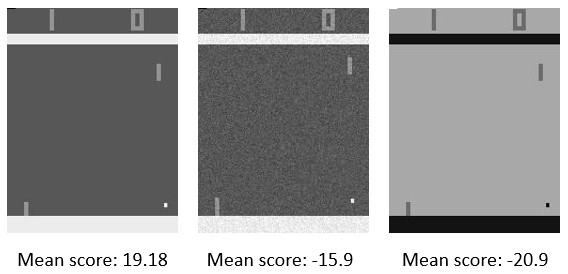} 
\caption{The agent are trained on Pong for 10 million frames, and test the agent on both Pong (left), Pong with Gaussian noise (middle) and invert color Pong (right). We use $\epsilon = 0.05$ for $\epsilon$-greedy policy and $\sigma = 15$ for noise, the mean scores are average of 10 episodes game play. }
\label{fig:4.1_worse_performance}
\end{figure}

Most deep reinforcement learning algorithm can achieve great performance in single environment, but agent often cannot handle a new \ichao{slightly altered} environment. 
In \figname~\ref{fig:4.1_worse_performance}, we show the screenshot of original Pong (the leftmost column) and two variations of Pong.
A DQN agent that pretrained on original Pong performs really bad on two of it's variations (-15.9 on Pong with Gaussion noise and -20.9 on Pong with inverted color).

Rusu~\etal~\cite{rusu2016progressive} analyzed the Pong to \ichao{noisy} Pong\ignore{ with noise} case, and found that the high-level filter on the clean task is not sufficiently tolerant to the added noise.
Thus some new low-level vision has to be relearned in order to adapt to the new task environment.
We found that by add\ichao{ing} some augmentations \ichao{to the source task during} training on the source task would help to avoid over-fitting and thus learn more key features of the task. 
We added a data augmentation layer before feed the input data into replay memory, the data augmentation layer will randomly transform the input. 
For example, Eq. \eqref{eq:4} demonstrates a data augmentation layer that will invert the color of environment state (screen) with the probability of 30\%, and remain unchanged otherwise. 

\begin{equation}
\label{eq:4}
S = \begin{cases} 1 - S & 30\% \\ 
S & \mbox{otherwise }\end{cases}
\end{equation}

\begin{figure}[ht]
\centering
\includegraphics[width=\linewidth]{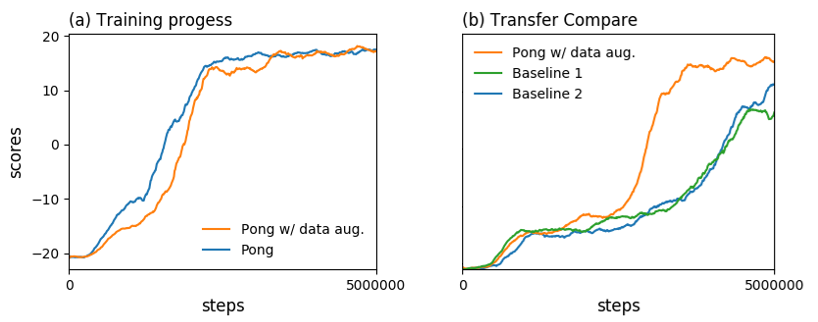} 
\caption{(a) The training progress of Pong and Pong with augmentation using Eq.\eqref{eq:4}.
(b) Training progress on a new environment using different pre-trained model as initial. The environment in (b) is Pong with Gaussian noise and $\sigma=50$. We use constant $\epsilon=0.1$ for $\epsilon$-greedy policy.}
\label{fig:4.1_toy_example}
\end{figure}

While this method increases the difficulty of training the agent, but with proper augmentation setting, the difficulty of training with augmentation is almost \ichao{the} same as \ichao{the origin} (\figname~\ref{fig:4.1_toy_example} (a)). 

In \figname~\ref{fig:4.1_toy_example}(b), we show three different \ignore{standard }DQN agent training progress on "Pong with Gaussian noise".
The ``\textit{Pong with data aug.}'' \ichao{stands for the agent trained by} using the pre-trained model of Pong with augmentation defined as Eq.\eqref{eq:4}. 
The \textit{Baseline 1} use\ichao{s} random initial, and \textit{Baseline 2} use\ichao{s} pre-trained model of Pong. 
It shows that using the pre-trained model of Pong with augmentation \ichao{obtains higher score within equal steps compared to} other \ichao{competitive methods}\ignore{baselines}.



\subsubsection{Detail Evaluation}
\label{subsubsec:augmentation-test}

We now further discuss the detail evaluation on augmentation setting, including why and how to find the good augmentation method for training. 

In \secname~\ref{subsec:augmentation}, we \ichao{prove} that \ichao{the agent trained on source task with augmentation} outperforms the alternative baseline transfer methods.
C. Zhang \etal~\cite{zhang2018study} have conduct a systematic study of standard reinforcement learning agents and find that they could overfit in various ways, and their result can support our finding which shows in \figname~\ref{fig:4.1_worse_performance} and \figname~\ref{fig:4.1_toy_example}.

\begin{figure}[ht]
\centering
\includegraphics[width=\linewidth]{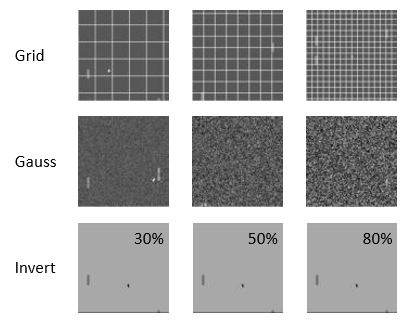} 
\caption{Different intensities of each augmentation on Pong. The first row shows the different size of grid noise, from left to right are 30, 20 and 10, the line color are all white; Second row is different level of Gaussian noise, from left to right are $\sigma=20$, $\sigma=50$ and $\sigma=80$; \ssarcandy{Third row is different frequency of inverted frames, the percentage means how often will the frame be inverted color.} }
\label{fig:4.2_aug-sample-frame}
\end{figure}

\ichao{Inspired by standard computer vision training pipeline, we added several varieties to the training set.}
By integrate augmentation method into reinforcement learning scenario, some constraint\ichao{s} must be added:
\begin{enumerate}
    \item The augmentation must not break the consistency of the visual content.
    \item 
    The augmentation should not be too hard for the agent to learn.
    \item 
    The augmentation should as different \ssarcandy{from original \ichao{visual content}} as possible to avoid over-fitting.
\end{enumerate}

The first constraint came from reinforcement learning characteristic, that the position on visual content is crucial, using augmentation method like rotate or flip would break the consistency of visual content. 
Second, using the augmentation that altered the \ichao{visual} content heavily would cause the agent fail to learn the task.
\ssarcandy{And for the third point, using visual altered augmentation \ichao{prevent the} agent \ichao{from} over-fitting and increase the ability of the agent to tolerate noise. 
It is important to find a balance between second and third point, \ie, the augmentation should be as different from original visual content as possible but not be too hard for the agent to learn.
}

We conduct an experiment on series of different augmentation setting, evaluate both the distance between with augmentation and without augmentation and the final performance on the task. 
We perform three sets of different augmentation method, including \textit{Gauss}, \textit{Grid} and \textit{Inverted}. 
Each method further test three different \ichao{variation levels}\ignore{intensities}, the sample frames of different \ichao{levels} of method \textit{Gauss} and \textit{Grid} \ichao{are shown}\ignore{shows} in \figname~\ref{fig:4.2_aug-sample-frame}, and the \textit{Invert} method \ichao{levels}\ignore{intensities} are defined as the invert frame \ichao{frequencies}\ignore{probabilities}.

\begin{figure}[ht]
\centering
\includegraphics[width=0.5\linewidth]{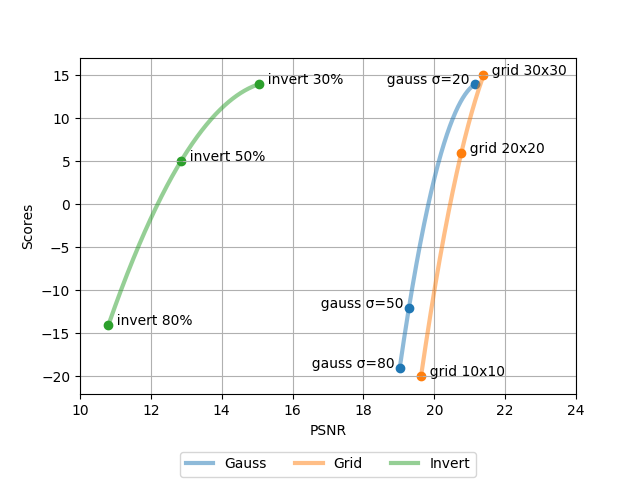} 
\caption{
Augmentation test for different augmentation method. We evaluate the difference and performance with different augmentation method on Pong, including \textit{Gauss}, \textit{Grid} and \textit{Inverted}. The x-axis means the difference between augmented Pong and normal Pong, which evaluated with PSNR, and the y-axis means the agent performance on normal Pong. The label beside data point indicated augmentation method and the intensity. \ssarcandy{We choose \textit{invert 30\%} for the augmentation method because it can not only remain high performance but also have farthest distance from normal Pong.}}
\label{fig:4.2_aug-test}
\end{figure}

The experiment results are shown in \figname~\ref{fig:4.2_aug-test}, we plot each augmentation method on a difference-performance 2D plot, where the x-axis represents \ichao{the appearance difference from} the augmentation \ichao{frame} to the normal one, and y-axis \ichao{indicates}\ignore{means that} how well an agent handle the task after training. 
We use PSNR to evaluate the \ichao{appearance} distance between \ichao{game playing frames}, the smaller PSNR means longer distance. 
We \ichao{made}\ignore{make} some observation\ichao{s} from this plot.
First, the stronger \ichao{level} of augmentation would cause the agent gradually unable to handle the task.\ignore{, and the PSNR value became smaller, which meet our expectation. }
We need to choose the one that meets the constraints describe above, that the augmentation should not be too hard for agent to learn but as different from normal as possible, although all three augmentations (\textit{Gauss}, \textit{Grid} and \textit{Inverted}) can have good performance in lower \ichao{level}\ignore{intensity} setting, i.e. \textbf{Gauss $\sigma=20$}, \textbf{Grid 30x30} and \textbf{Inverted 30\%}, but the \textbf{Inverted 30\%} is the only one that has smallest PSNR value and still remains good performance, so it would be the better augmentation setting.

\begin{figure*}[ht]
\centering
\includegraphics[width=\linewidth]{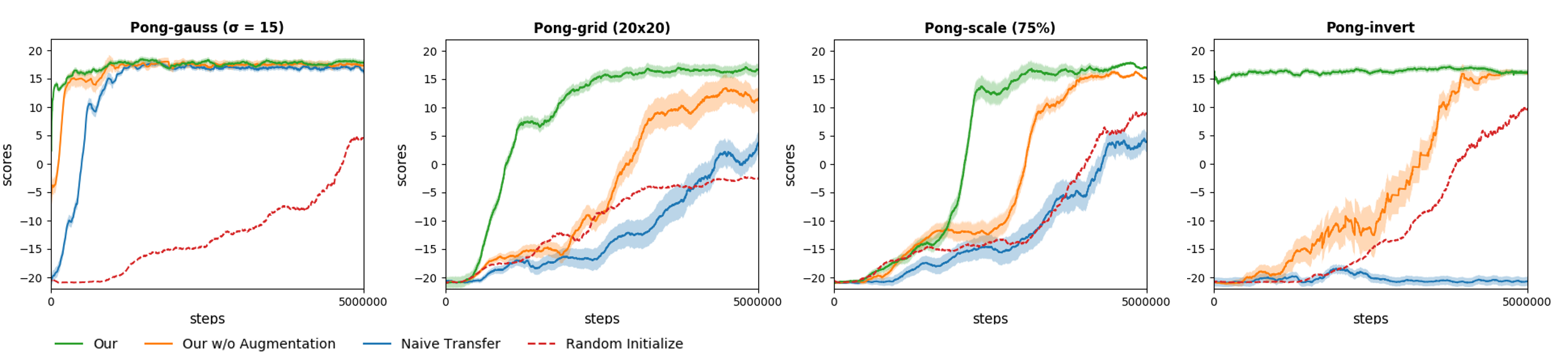} 
\caption{
Transfer progress of Pong variants. Pong variants include noisy, inverted color and scaled transforms. The results are averaged over 3 runs and the shadow represent standard deviation.}
\label{fig:5-1_pong_variants}
\end{figure*}

\section{Experiments}
\label{sec:experiments}
\begin{figure}[ht]
\centering
\includegraphics[width=\linewidth]{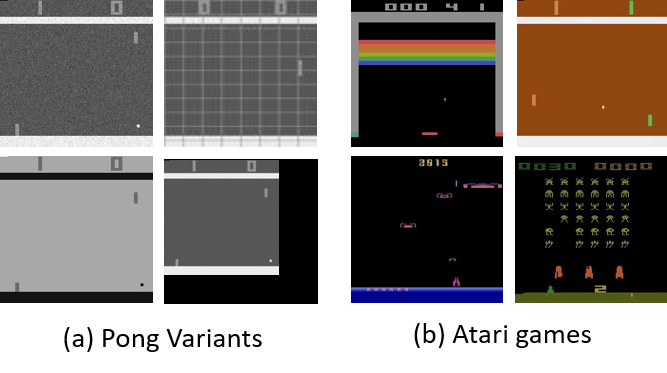} 
\caption{Example frames from different domain\ichao{s}. (a) Pong variants: noisy, recoloured and scale transforms; (b) Atari games.
\ssarcandy{From top right to bottom left: Breakout, Pong, Demon Attack, Space Invader}.
These games introduce a more challenging setting for transfer.}
\label{fig:5_example_frames}
\end{figure}

In the following experiments, we evaluate \ichao{the transfer effectiveness of} our method \ichao{using the Arcade Learning Environment (ALE)~\cite{bellemare13arcade}}.
\ichao{First, we conducted an experiment to evaluate how our method improve the transfer effectiveness across synthetic variations of Pong}.
\ichao{Second}, we experiment on \ichao{a} more challenging setting\ichao{, \ie~}transfer between different Atari games. 
We show sample frames of the selected tasks in \figname~\ref{fig:5_example_frames}.
\ichao{
In each experiment, we compare the agent performances of following four methods:
\begin{itemize}
    \item \textit{random baseline}: train DQN directly target task with random initialized weights.
    \item \textit{na\"{\i}ve baseline}: train DQN on target task with the pre-trained weights on source task.
    well-trained source task model parameters to target task network
    \item \textit{our method w/o augmentation}: DQN with adversarial objective trnasfer (described in \secname~\ref{subsec:transfer_learning}).
    \item \textit{our full method}: DQN with adversarial objective transfer and frame augmentations.
\end{itemize}
}

\subsection{Pong Variants}
\label{subsec:pong_variants}
The first evaluation domain is a set of synthetic variants of the Atari game ``Pong''. 
\ichao{We created synthetic variants by altering visual appearances of the original ``Pong''.}
The variants of Pong are \textbf{Noisy} (Gaussian noise is added to the inputs), \textbf{Grid} (fixed grid lines are added on input), \textbf{Invert} (input color is inverted), and \textbf{Scale} (input is scaled by 75\% and with black padding). 
Example frames are shown in \figname~\ref{fig:5_example_frames} (a).
\ichao{The purpose of this experiment is to test}\ignore{thus providing a setting where we \ichao{are}\ignore{can be} confident that} \ichao{whether the agent can learn the core game play across it's visual variants}.

\figname~\ref{fig:5-1_pong_variants} shows the transfer progress \ichao{of}\ignore{in} each variant.
\ssarcandy{Overall, \ichao{our method (both with and without augmentation versions)}\ignore{ of our method} achieve a better learning progress on target task compare to baseline methods. 
\ichao{Our method saves} at least\ignore{up to} 1 millions frames of training time to reach the convergence state for \textbf{Pong-grid}, \textbf{Pong-scale} and \textbf{Pong-invert}.
}

\begin{figure}[ht]
\centering
\includegraphics[width=\linewidth]{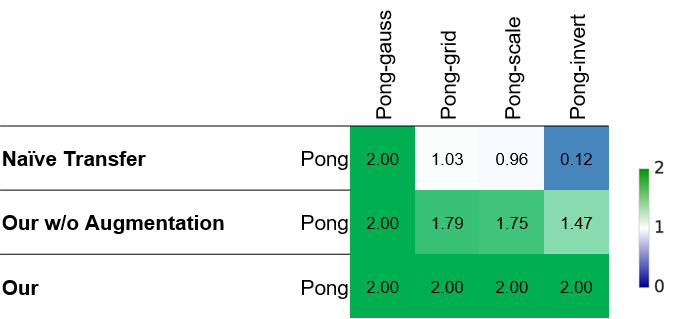} 
\caption{Transfer score matrix. The higher score means the better transfer performance. Colours indicate transfer scores (clipped at 2). }
\label{fig:5-2_pong_transfer_score}
\end{figure}

\ichao{Moreover}\ignore{For a more clearly look}, we measure the transfer performance by the transfer score~\cite{rusu2016progressive}.
\ichao{The transfer score is defined as the relative performance of an method compared with \textit{random baseline} method.}
\ssarcandy{
Higher transfer score means more effective transfer. 
Transfer score larger than 1 means the performance is relatively better than \textit{random baseline} method, \ichao{\ie~positive transfer.}
\ichao{On the contrary, transfer score smaller than 1 means the performance worse than \textit{random baseline} method, \ie~negative transfer.}
}

As shown in \figname~\ref{fig:5-2_pong_transfer_score}, 
we can \ichao{observe} that our method get better transfer scores \ichao{across} all experiments. 
Interestingly, the \textit{naive baseline} (initialized with pre-trained weights on ``Pong'') obtains \ichao{worst} transfer score\ignore{ are not pretty good}.
\ichao{It} indicates that there \ichao{are}\ignore{would have} negative transfer effect\ichao{s}, especially for \textbf{Pong-invert} case.

Our method \ichao{boosts the}\ignore{provided an increase in } learning speed \ichao{for both with and without augmentation versions of our method.}
\ignore{used are much better than the Naive Baseline approach.} 
\ichao{As shown in \figname~\ref{fig:5-1_pong_variants}, we can observe that} our method with augmentation \ichao{learns}\ignore{, the learning speed gets a } significant \ichao{faster}\ignore{ improved} compared to the version without augmentation.
\ichao{This shows} that \ichao{our method with augmentation can significantly help agent to learn the core game play across visual variants of a single game.}

We further look close\ichao{ly} at specified \ichao{transfer pair} cases.
For \textbf{Pong-gauss} case, the difference between source and target \ichao{task}\ignore{environment} is \ichao{smallest}\ignore{shortest}.
\ichao{We can observe that even the \textit{na\"{\i}ve baseline} method provide\ichao{s} a very good transfer effect}.
Although \ichao{the \textit{na\"{\i}ve baseline} method reaches max} transfer scores (\ie~2.0), our method still \ichao{outperform}\ignore{beat} it on coverage time as shown in \figname~\ref{fig:5-1_pong_variants}. 
For \textbf{Pong-invert} case, it shows that \ichao{\textit{na\"{\i}ve baseline} method} fails to learn on target task, which means that the \ichao{knowledge learned from source task}\ignore{source task parameters}
are \ichao{hindering the agent to learn target task.}
\ichao{O}n the other hands, our method without augmentation \ignore{can}minimize\ichao{s} the negative effect because the generator will try to produce features that \ichao{are} similar with source task's features.
In terms of our \ichao{full} method, it achieves great performance at starting time is because that the source task is trained with 30\% inverted frame, in other words, the source agent has learned \textbf{Pong-invert} at training time, thus it can handle this specified case well. 
In general speaking, our method obtains the best \ichao{transfer} result, followed by our method without augmentation, and \ichao{na\"{\i}ve transfer} method obtains worst result.


\subsection{Multi-level Transfer}
\label{subsec:cross_level}
Next, we designed a multi-level game scenario to test \ichao{the effetiveness of} our transfer method.
It is common that for each game, the designer designed multi-level\ichao{s with increasing difficuties.}
Instead of training the agents separately on each level, \ichao{we} use \ichao{the proposed} transfer learning method \ichao{to increase the learning speed when facing harder levels.}

\begin{figure}[ht]
\centering
\includegraphics[width=\linewidth]{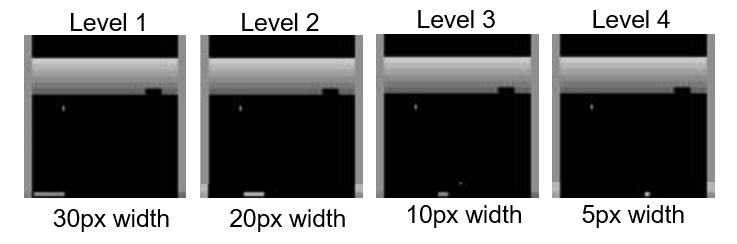}
\caption{
Sample frames of different level of Breakout.
}
\label{fig:5-2_cross_level_example_frames}
\end{figure}

In our scenario, we build a \ichao{multi-level version of} Breakout\footnote{Environment source code can be found at https://github.com/SSARCandy/breakout-env} with different difficulties.
We designed 4 different levels \ichao{with increasing difficulties} by shrinking the paddle width, the different paddle widths are: 30px width (level-1), 20px width (level-2), 10px width (level-3) and 5px width (level-4).
\figname~\ref{fig:5-2_cross_level_example_frames} show 4 different levels sample frames.

\begin{figure}[ht]
\centering
\includegraphics[width=\linewidth]{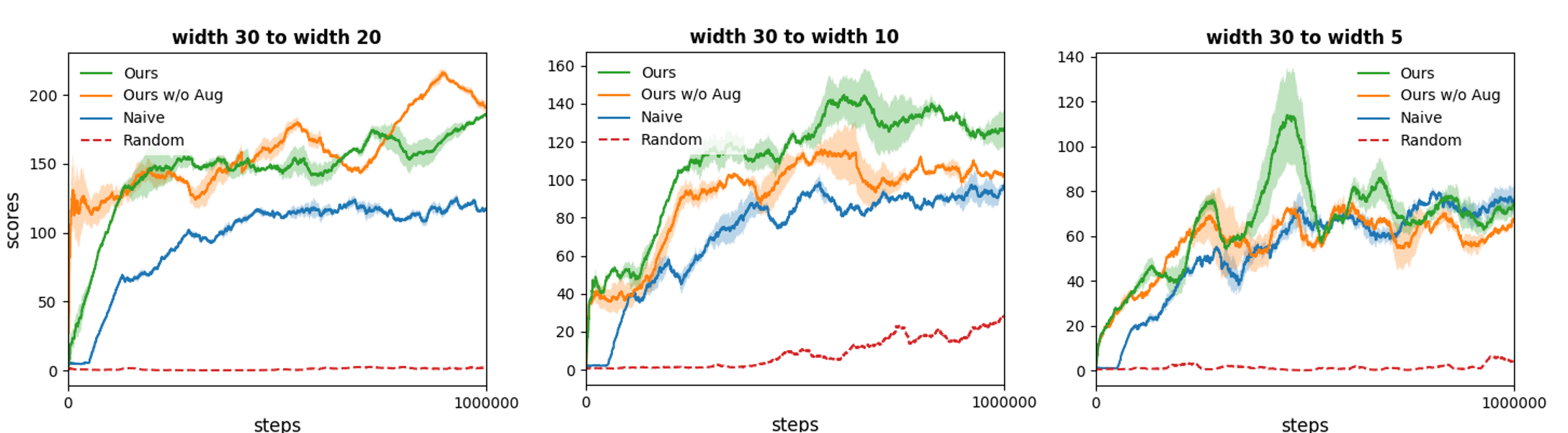}
\caption{
Transfer progress of cross-level scenario.
}
\label{fig:5-2_cross_level_result}
\end{figure}

And we perform 3 transfer cases, including level-1 to level-2, level-1 to level-3 and level-1 to level-4.
The training progress shown as \figname~\ref{fig:5-2_cross_level_result}.

We can find that the worst performance is ``Random initialize'' (red dashed line) because \ichao{the agent was trained} from scratch, without any prior knowledge.
And all other transfer methods have very good transfer effectiveness and can be reached stable performance within 1 million steps.
For width 30 to width 20 (level-1 to level-2) case, the performance of our methods (both w/ and w/o augmentation) are better than the Naive method, and augmentation (orange line) helps \ichao{the} agent to learn fast\ichao{er} at the beginning.
For width 30 to width 10 (level-1 to level-3) case, our methods still have a noticeable \ichao{boost} of learning effectiveness than Naive method.
Finally, for with 30 to width 5 (level-1 to level-4) case, there is barely any difference between Navie method and our methods.
This suggested that the difficulties of the final level (\ie~level 4) has significant different from all the previous levels.
The performance of the transfer method can then be used to identify the difficulty differences during the game designing process.

\subsection{Cross Games Transfer}
\label{subsec:atari_game}
Next we investigate the transfer \ichao{effectiveness} between different Atari games. 
\ichao{W}e select 4 different games from Atari 2600 to perform cross game transfer experiments, including Pong, Breakout, SpaceInvader, and DemonAttack. 
\ichao{We choose} Pong and Breakout \ichao{because they share}\ignore{are considered as having} some similarit\ichao{ies of gameplay, \ie~the player}\ignore{ because both gameplay consists of } tr\ichao{ies} to hit a moving ball with a rectangular paddle. 
\ichao{On the other hand,}\ignore{And} SpaceInvader \ichao{and}\ignore{is considered similar to } DemonAttack \ichao{shares similar gameplay, \ie~the player}
need to shoot moving enemies. 
In this experiments, we perform transfer between these selected games.


\begin{figure}[ht]
\centering
\includegraphics[width=\linewidth]{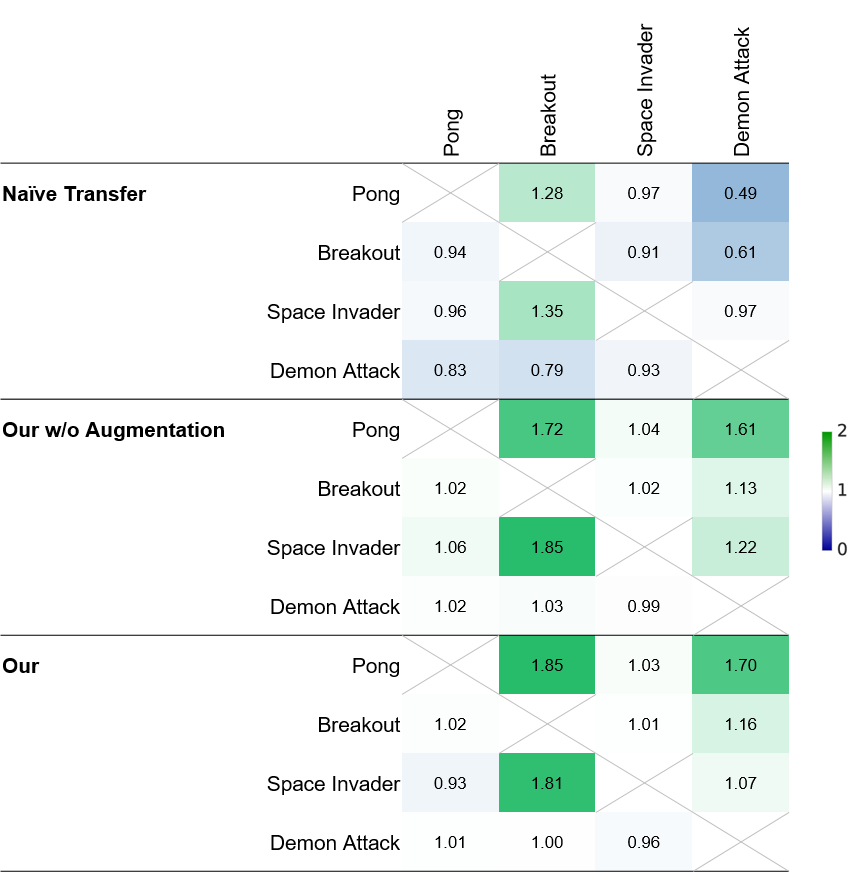} 
\caption{Cross game transfer score matrix. 
The higher score \ichao{indicates}\ignore{means} the better transfer performance. 
Colours \ichao{encode}\ignore{indicate} transfer scores (clipped at 2). 
The X sign mean\ichao{s} source and target task are \ichao{the} same and no need to transfer.
}
\label{fig:5-3_cross_game_transfer_score}
\end{figure}

\ichao{We}\ignore{The  summary} \ichao{report} the transfer results\ignore{are reported as} in the transfer scores matrix shown in \figname~\ref{fig:5-3_cross_game_transfer_score}.
In this matrix, we compare three different \ichao{methods}\ignore{approaches}, including na\"{\i}ive \ichao{transfer} baseline, our \ichao{method} w/o augmentation, and our \ichao{full} method. 

Overall, the \ichao{na\"{\i}ve transfer} method often got transfer score that \ichao{is} smaller than 1, \ichao{which indicates}\ignore{means} that it needs more training time compare to training on target task directly, thus have negative transfer effect. 
\ignore{For }\ichao{O}ur approaches \ichao{obtain higher} transfer scores than \ichao{na\"{\i}ve transfer} approach in most of the cases, \ichao{which indicates that our methods} help to learn the target tasks.
And in some cases, although transfer score\ichao{s are}\ignore{is} not high, our method eliminates the negative effect \ichao{that introduced by na\"{\i}ve transfer method}.

For our method with augmentation, the benefit \ichao{of} augmentation is not so significant \ichao{as}\ignore{like our experiments} in Pong variants \ichao{experiment} described in \secname~\ref{subsec:pong_variants}.
In some case\ichao{s} like \textbf{Pong to Breakout} and \textbf{Pong to DemonAttack}, augmentation still helps the \ichao{agent}\ignore{model} learn\ichao{s} faster\ignore{ than without it}. 
And in other cases, the performance\ichao{s} with and without augmentation did not \ichao{demonstrate}\ignore{shows a} noticeable \ichao{differences}\ignore{benefit}.

\ichao{We also observed that some of the transfer performances of the similar gameplay pairs} 
(Pong and Breakout, SpaceInvader and DemonAttack) \ichao{are not obviously improved.}
We believe that this is because these Atari games are too different than some similarity between games could not take advantage when transferring.



    

\section{Conclusion}
\label{sec:conclusion}
In this works, we investigate the knowledge transfer for deep reinforcement learning\ignore{scenario}. 
Unlike previous works \cite{rusu2016progressive,parisotto2015actor} that \ichao{requires training multiple agents on}\ignore{need} multiple source task\ichao{s} for generaliz\ichao{ing} and transfer\ichao{ing} to target task, we proposed a method that can accelerate the training progress on a new task with only single prior task.\ignore{ knowledge.} 
Furthermore, we found that \ichao{with}\ignore{ using} a \ignore{deadly}simple \ichao{data} augmentation method\ichao{,} \ichao{the agent can learn the}\ignore{can help} target task \ichao{faster}\ignore{have a better initial guess of model parameters}. 
And we \ichao{demonstrated}\ignore{proved that} our \ichao{method}\ignore{approach} outperforms baselines in both easy and challenging cases \ichao{using Atari 2600 benchmark.}\ignore{by evaluating on popular benchmark Atari 2600 domain.}

\bibliographystyle{IEEEtran}
\bibliography{references}
\end{document}